\documentclass[letterpaper, 10 pt, conference]{ieeeconf}  % Comment this line out if you need a4paper

\IEEEoverridecommandlockouts                              % This command is only needed if 
\usepackage{graphics} % for pdf, bitmapped graphics files
\usepackage{graphicx}
\usepackage{float} %设置图片浮动位置的宏包
\usepackage{subfigure} %插入多图时用子图显示的宏包
\usepackage{amsmath}
\usepackage{mathptmx} % assumes new font selection scheme installed
\usepackage{hhline}
\usepackage{underscore}
\usepackage{amssymb}
\usepackage{float}
\usepackage{stfloats}
\usepackage{pdfpages}
\usepackage{caption}
\usepackage{hyperref}
\hypersetup{hidelinks,
	colorlinks=true,
	allcolors=blue,
	pdfstartview=Fit,
	breaklinks=true}

\title{\LARGE \bf
RWT-SLAM: Robust Visual SLAM for Highly Weak-textured Environments
}

\author{Qihao Peng, Zhiyu Xiang, YuanGang Fan, Tengqi Zhao, Xijun Zhao% <-this % stops a space
\thanks{Qihao Peng, Zhiyu Xiang, YuanGang Fan and Tengqi Zhao are with the Dept of Information and Electronic Engineering, Zhejiang University, Hangzhou, 310027, China. Zhiyu Xiang is the corresponding author, xiangzy@zju.edu.cn. Xijun Zhao is with China North Vehicle Research Institute, Beijing, 100072, China.}% <-this % stops a space
\thanks{This work is supported by NSFC-Zhejiang Joint Fund for the Intergration of Industrialization and Informatization under grant No.U1709214 and the Key Research \& Development Plan of Zhejiang Province (2021C01196).}%
}

\begin{document}
\maketitle
\begin{abstract}
As a fundamental task for intelligent robots, visual SLAM has made great progress over the past decades. However, robust SLAM under highly weak-textured environments still remains very challenging. In this paper, we propose a novel visual SLAM system named RWT-SLAM to tackle this problem. We modify LoFTR network which is able to produce dense point matching under low-textured scenes to generate feature descriptors. To integrate the new features into the popular ORB-SLAM framework, we develop feature masks to filter out the unreliable features and employ KNN strategy to strengthen the matching robustness. We also retrained visual vocabulary upon new descriptors for efficient loop closing. The resulting RWT-SLAM is tested in various public datasets such as TUM and OpenLORIS, as well as our own data. The results shows very promising performance under highly weak- textured environments.
\end{abstract}

%%%%%%%%%%%%%%%%%%%%%%%%%%%%%%%%%%%%%%%%%%%%%%%%%%%%%%%%%%%%%%%%%%%%%%%%%%%%%%%

\section{INTRODUCTION}

Visual simultaneous localization and mapping (SLAM) is an essential task for mobile robots navigation as well as in AR/VR. Nowadays feature based SLAM algorithms are popular for their high performances and computational efficiency. However, robust SLAM under highly weak-textured environments still remains a challenging problem. Hand-crafted features such as SIFT [1], ORB [2] and Shi-Tomas [3] are not able to extract reliable keypoints for matching in regions with highly low texture or motion blur. In fact, these extreme environments are difficult for whatever the state-of-art feature-based methods [4] or direct methods [5][6]. For some textureless but well structured environment,  some solutions are proposed to strengthen the feature matching by integrating line or plane features, e.g. Stereo-PLSLAM [7], SuperLine [8] and Structure-SLAM [9]. However, these methods rely heavily on visible structures like object edges and planes. In environments with little structure and texture, tracking may still fail.

On the other hand, there is a trend to replace hand-crafted features with deep features to improve the robustness of the SLAM systems. Trained with large mount of diversified data, the deep learning-based methods can operate on full-size image and jointly compute pixel-wise interest points and the associated descriptors. Experiments in [10] indicate that SuperPoint can produce more distinctive descriptors than classical methods and the interest point detector is on par with hand-crafted features. Similarly, GCN [11] uses a recurrent neural network to predict the location of keypoints as well as their descriptions for camera motion estimation. GCNv2 [12] simplifies the network for efficiency and incorporates the learned features into ORB-SLAM2 [4] to form GCN-SLAM. Although these deep learning-based methods perform better than the traditional ones in complex environments, when dealing with highly weak-textured scenes, there is still much room for improvement.
\begin{figure}[tpb]	
\centering	
	\includegraphics[scale=0.28]{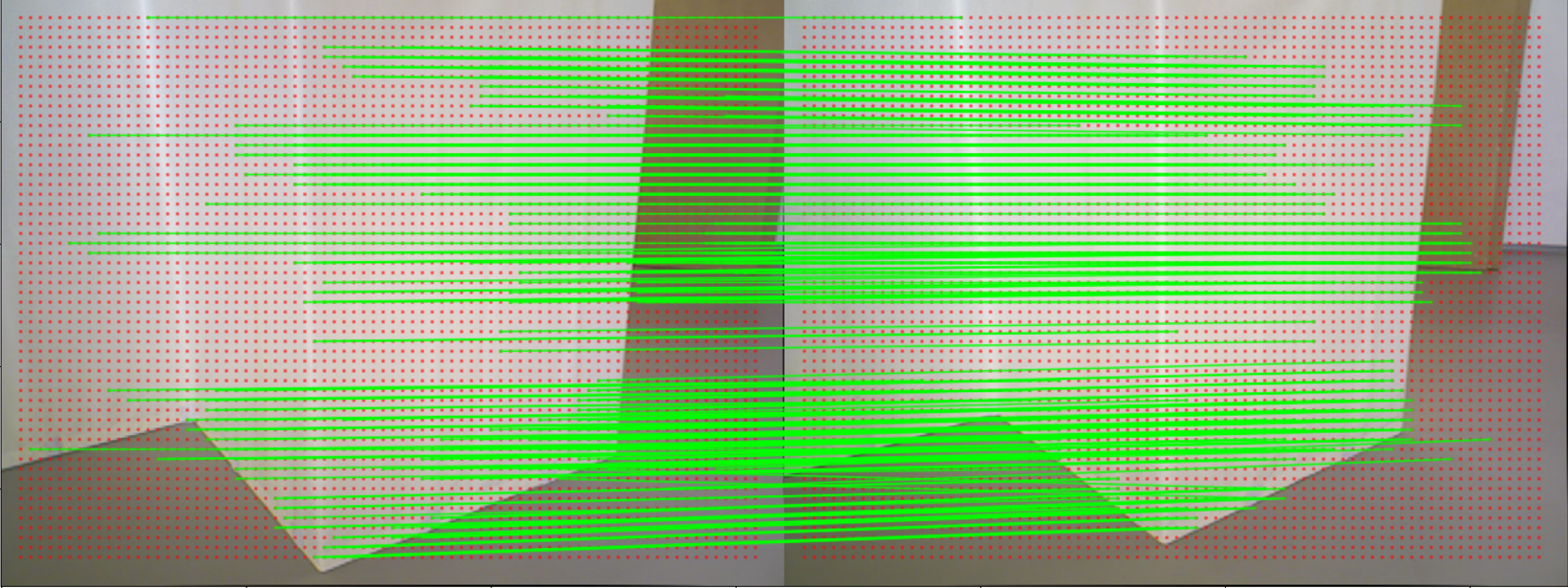}
		\label{pretrain(a)}
		\vspace{1mm} %调整纵向距离
	\includegraphics[scale=0.28]{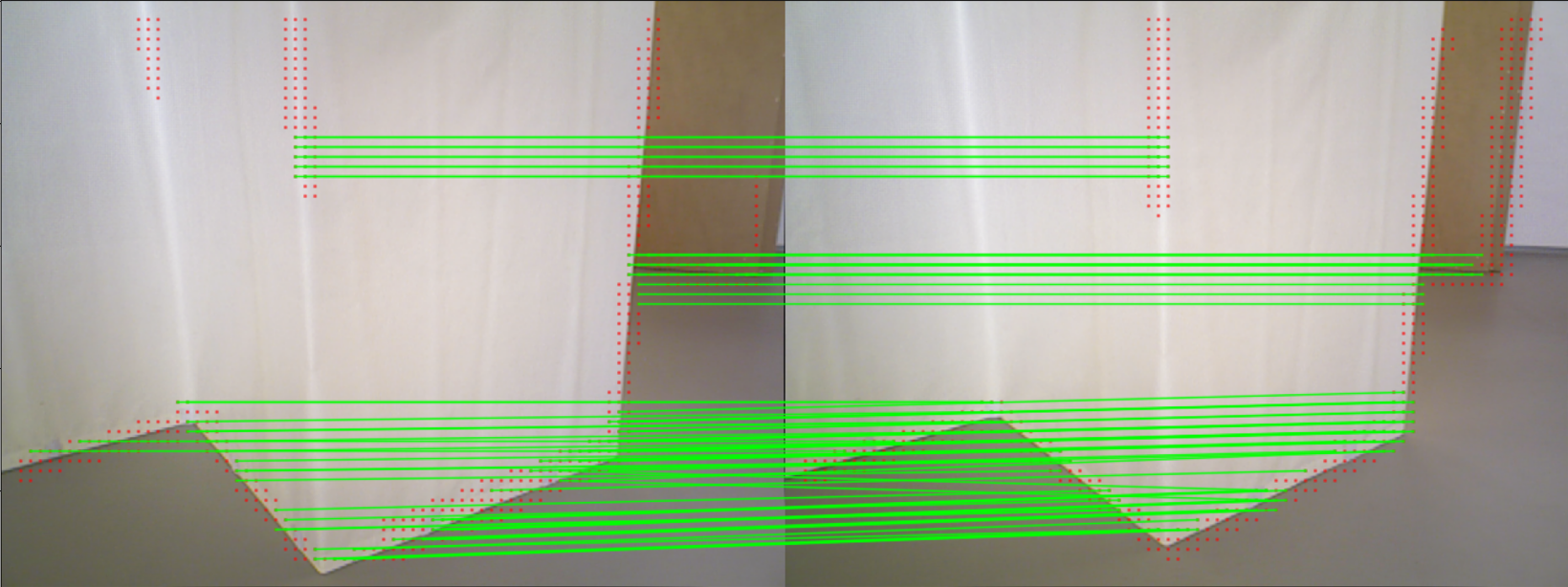}
		\label{pretrain(b)}
\caption{A matching example in areas of weak-texture using keypoints and descriptors of our system. The point correspondence in top and bottom images are produced by KNN matcher before and after applying the feature mask, respectively.}	
% \begin{center}
% Figure 1. This is the caption defined by myself.
% \end{center}
\label{pretrain}
\end{figure}

Recently, several works [13][14] propose to directly predict pixel-wise matches for a pair of images by using CNN. LoFTR [14] utilizes transformer [15] with self-attention and cross-attention mechanisms to produce dense matches. They are able to produce large quantities of matches under low textured environments without using any feature descriptor. This motivates us to induce this state-of-the-art method into visual SLAM. However, when purpose of accurate localization of SLAM is concerned, the detector-free matching becomes a drawback. The reason lies in three aspects. Firstly, although LoFTR can produce dense matches for a pair of images, the repetitiveness of the matching pairs across the neighboring images is not guaranteed. This makes it difficult for further optimization of the features’ position across multiple frames, which modern SLAM systems high rely on. Secondly, popular SLAM systems like ORB-SLAM2 [4] rely on feature descriptors to match the feature points to the local map, and use PnP algorithm to produce good initial estimation for poses. Lack of feature descriptors makes this stage infeasible. Lastly, despite the dense matching, the pixel positions produced by the detector-free methods are not as accurate as the features extracted from those detector based methods such as ORB. In particular, lots of matching outliers can be observed in highly low textured regions. 

In this paper, we propose a novel visual SLAM system named RWT-SLAM, which corresponds to very \textbf{R}obust SLAM in \textbf{W}eak-\textbf{T}extured environments. We modify LoFTR network in order to produce distinctive feature descriptors. To integrate the new features into the popular ORB-SLAM framework, we develop feature masks to filter out the unreliable matches and employ KNN to strengthen the robustness of the matching. We also trained visual vocabulary upon new descriptors for loop closing. The resulting RWT-SLAM is tested in various highly weak-textured environments and shows very promising performance. The contributions of this paper are summarized as follows:
\begin{itemize}
\item A novel full visual SLAM system capable of robustly working under highly weak-textured environments is proposed. To the best of our knowledge, it is the first SLAM algorithm which can achieve successful tracking on all sequences with either \textit{no structure or no texture} labels in TUM dataset [35].
\item We demonstrate that a detector-free network aimimg at producing dense feature matches can be reformed for SLAM applications. We extract coarse and fine level deep features from LoFTR [14] to construct the final descriptors. 
\item Comprehensive experiments are carried out on public TUM RGB-D [35], OpenLORIS [36] and our own dataset, demonstrating the superior localization and robustness performance under extremely weak-textured environments.
\end{itemize}

\section{RELATED WORKS}

\subsection{Traditional Visual SLAM}

Traditional visual SLAM algorithms can be roughly divided into two classes: direct (photometric-based) methods and indirect (feature-based) methods. The direct approaches estimate motion from phtotometric changes of the image while indirect methods rely on a subset of features of the image. LSD-SLAM [5], a direct monocular method, can build semi-dense consistent maps for large-scale scenes. The sparse direct visual odometry DSO [6] omits the smoothness prior used in other direct methods and instead samples pixels evenly throughout the images. Most indirect methods rely on PTAM [16], which is the first algorithm splitting the tracking and mapping thread separately to achieve better performance. The popular ORB-SLAM [17] employs ORB [2] features and incorporates three threads, i.e., tracking, local mapping and loop closing to fullfill the task. In particular, SVO [18] is a semi-direct odometry extracting sparse features and operating directly on pixel intensities around the features. Comparing with the feature-based methods, indirect approaches perform better on images with low or repetitive textures. However, they are much more sensitive to illumination changes of the environment and more difficult to initialization.

\subsection{Deep learning based SLAM}

Many deep learning based SLAM methods are proposed in recent years. They are usually formed by replacing one or more modules of the traditional SLAM framework with deep networks [19][20][21][12][22]. To improve depth predictions that are used to initialize the SLAM system, CNN-SLAM [19] incorporates a depth estimation network within the popular LSD-SLAM framework to produce dense scene reconstructions with metric scale. DS-SLAM [21] utilizes a semantic segmentation network to filter out features on moving objects, achieving better localization accuracy in highly dynamic environments. The most similar methods to ours are GCN-SLAM [12] and DXSLAM [22]. In GCN-SLAM [12], keypoints and binary descriptors produced by the neuro-network named GCNv2 is employed to replace ORB [2] features used in ORB-SLAM2 [4]. DXSLAM [22] uses keypoints and descriptors generated by a pretrained HF-Net to improve system's performance. Compared with GCN-SLAM and DXSLAM, we obtain the features from a modified transformer based feature matching network LoFTR and integrate them into the ORB-SLAM framework. The resulting algorithm can achieve much higher performance than its counterparts under highly weak-textured environments. 

Other works have attempted to train end-to-end SLAM systems [23][24]. Most of them are not full SLAM systems, but focus on small scale reconstruction on several frames. They do not have key modules of the modern SLAM system such as loop closure and global bundle adjustment, which limits the accuracy of the system. Recently, DeepSLAM [25] integrate three sub-network to imitate full modules of SLAM. However, they need to be trained for each benchmark from scratch, limiting the generalization ability for practical use.

\subsection{Deep Feature Matching}
The deep feature matching methods can be roughly divided into two categories. The first one focuses on learning to produce keypoints and their descriptors simultaneously [10][26][27][28][29]. SuperPoint [10] proposes a self-supervised strategy to train a fully convolutional network for joint keypoints detection and description. In D2-Net [26], a local maxima within and across feature maps are selected as interest points and the descriptors are generated from the same feature maps. R2D2 [27] utilizes a predictor of discriminativeness to avoid ambiguous areas.

The second category is detector-free methods which directly learn dense matches or descriptors without explicit keypoint detection phase [30][31][32][13][14]. NCNet [30] uses 4D cost volumes to enumerate all possible matches between a pair of images. More recently, DRC-Net[31] follows this line and trains a CNN in a coarse-to-fine manner with synthetic transformations. In [32], a novel weakly-supervised framework is proposed by using solely relative poses between images to learn descriptors. Inspired by SuperGlue [33] which is a detector-based feature matching method, LoFTR [14] proposes a transformer based detector-free design to produce dense matches for various complex environments. It is able to produce large quantities of pixel level matches even under highly weak structured or textured environments. 

\begin{figure*}[tpb]	
\centering	
	\includegraphics[width=1.0\linewidth, height=0.35\linewidth]{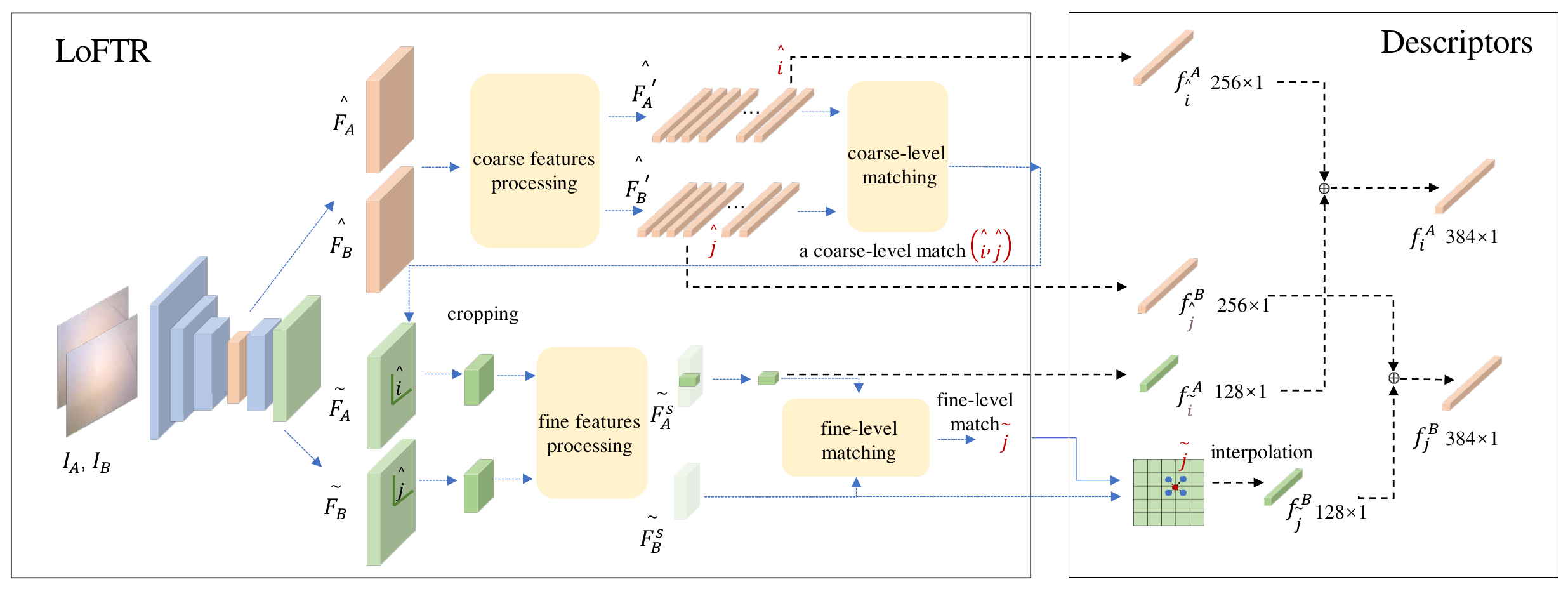}
		\label{network}
\caption{The keypoint and descriptor generation pipeline. Coarse-level and fine-level feature maps are generated within the LoFTR, where we further extract the coarse and fine level descriptors and concatenate them together to construct the final descriptors for the feature points.}	
\end{figure*}

\section{METHOD}

In this section, we introduce our RWT-SLAM algorithm in detail. The generation of keypoint descriptors from the detector-free network LoFTR [14] is introduced in part A. Then the reformation of ORB-SLAM2 [4] to make it adaptive to the new feature points is described in part B. 

\subsection{Feature Descriptors} 
The descriptor generation pipeline is illustrated in Fig.2. Given a pair of images $I_{A}$ and $I_{B}$, LoFTR [14] establishes coarse-level matches with coarse features and then refines the matches to the fine scale with a coarse-to-fine scheme.

\textbf{Coarse-level descriptors:} As suggested in Fig. 2, the coarse features $\hat{F_{A}}$ and $\hat{F_{B}}$ are from the fourth blocks of the CNN with 256 channels with $1/8$ scale of the original image. They are fed into a transformer with attentional aggregation and receptive filed expansion for further feature extraction. Then a matching module is utilized to construct coarse-level matches with an outlier rejection. Based on the transformed features $\hat{F_{A}^{'}}$ and $\hat{F_{B}^{'}}$, the score matrix S whose elements $S_{\hat{i},\hat{j}}$ can be computed by the inner product of corresponding feature vectors $f_{\hat{i}}^{A}$ and $f_{\hat{j}}^{B}$ with

\begin{equation}
S_{\hat{i},\hat{j}}=f_{\hat{i}}^{A}\cdot f_{\hat{j}}^{B}, \quad where \quad (f_{\hat{i}}^{A},f_{\hat{j}}^{B})\in(\hat{F_{A}^{'}},\hat{F_{B}^{'}})
\end{equation}

Inspired by the matching strategy, we notice that the transformed coarse features can be utilized to produce descriptors for a match $(\hat{i},\hat{j})$. Therefore, the corresponding items on the coarse feature maps $\hat{F_{A}^{'}}$ and $\hat{F_{B}^{'}}$ are separately saved as the coarse-level descriptors, namely $f_{\hat{i}}^{A},f_{\hat{j}}^{B} \in \mathbb{R}^{256}$. Thanks to the self-attention and cross-attention mechanisms of the transformer module, the coarse-level descriptors encode the unique description of a local 8×8 image patch while maintaining global receptive filed of the whole image.

\textbf{Fine-level descriptors:} LoFTR [14] operates on fine-level features to obtain sub-pixel matches in a coarse-to-fine manner. The fine-level features are cropped within a local window of size $5\times5$ centered at $\hat{i}$ and $\hat{j}$ respectively. Then a fine-level processing and matching pipeline is applied to get a sub-pixel coordinate $\tilde{j}$ on the second image $I_{B}$. The final matched positions $(i,j)$ can be formulated as:
\begin{equation}
\begin{gathered}
i=8\times \hat{i}\\
j=8\times \hat{j}+2\times \tilde{j},
\end{gathered}
\end{equation}
We select the center feature vector of $\tilde{F^{s}_{A}}$ as the fine-level descriptor on image $I_{A}$ for position $i$. The fine-level descriptor on image $I_{B}$ is obtained by interpolating $\tilde{F^{s}_{B}}$ bilinearly at position $\tilde{j}$, as shown in the right part of Fig. 2. At this point, we have obtained the fine-level descriptors $f_{\tilde{i}}^{A},f_{\tilde{j}}^{B} \in \mathbb{R}^{128}$. Finally, by concatenating the coarse and fine level descriptors, the complete hierarchical descriptors that capture both high and low level textures of the matched feature $(i,j)$ can be obtained and recorded as:

\begin{equation}
\begin{aligned}
f_{i}^{A}=f_{\hat{i}}^{A} \oplus f_{\tilde{i}}^{A}  \qquad \in \mathbb{R}^{384}\\
f_{j}^{B}=f_{\hat{j}}^{B} \oplus f_{\tilde{j}}^{B} \qquad \in \mathbb{R}^{384}
\end{aligned}
\end{equation}

\subsection{Framework of RWT-SLAM}
The system overview of our RWT-SLAM is illustrated in Fig. 3. Given the current frame, the keypoints and descriptors from the reformed LoFTR are fed into the classical ORB-SLAM framework. To make our RWT-SLAM more adaptive to the extreme environment, we make more improvements on the  ORB-SLAM framework.

\textbf{Feature extraction and filtering.} LoFTR [14] directly produces dense feature matches for a pair of images. However, the distribution of the matches can change drastically across images, especially for the scene with large illumination changes or weak texture. When applying them on successive input frames with weak texture, the repetitiveness of the feature points across the frames can be low, which is fatal for SLAM. To produce enough feature points with wide distribution, we copy the current frame and concatenate them at the channel dimension before feeding them into LoFTR. Generally for each $8\times8$ image patch, one feature point and corresponding descriptor can be obtained.

The feature points we obtained have sufficient quantity while at the cost of less distinctiveness. If there is some structure in the environment, which is true for common corridors or offices, the structure information can be used to select more distinctive feature points. We rely on canny edge detection followed by Hough transformation to form feature masks with the shape of thick line. Only the feature points within the mask are maintained for matching because they are more distinctive and are able to produce more accurate matching. An example showing the matching results before and after the use of feature mask is presented in Fig. 1. Some examples of the feature masks are also shown in Fig. 4, where the width of the line is set to 20 pixels. 

\textbf{Tracking.} In ORB-SLAM2 [4], the motion estimation is achieved by frame to frame keypoints tracking and pose-only bundle adjustment. Once the keypoints and the descriptors are obtained, the system relies on a progressive methods for frame to frame tracking. First, by assuming a constant velocity the keypoints of the previous frame are projected to the current frame and matches are searched in a local window. If that fails, the system will try to find matches using bag of words (BoW) among current frame and the referenced keyframes. In our system, we replace these two matching strategies with a standard K nearest-neighbor search followed by a ratio test. The KNN matcher establishes stable matches by computing the Euclidean distance between the descriptors. With the learned descriptors, the brute-force matcher is more suitable for images with low-texture. 

\textbf{Vocabulary traning.} A visual vocabulary is trained offline in ORB-SLAM2 [4] to accelerate matching process in tracking, relocalization and loop closing. We adopt DBoW3 framework to build the new visual vocabulary based on the descriptors produced in part A, and Bovisa 2008-09-01 [34] is used for vocabulary training. The vocabulary is produced in binary form, which is more efficient for use during system initialization and image matching.

\section{EXPERIMENTS}
We carried out extensive experiments on various datasets to verify the effectiveness of our SLAM system. Similar to GCN-SLAM and DXSLAM, all experiments are conducted on RGB-D data. The TUM RGB-D dataset [35] contains sequences with highly low texture, which are very challenging to most existing visual SLAM algorithms. We further perform experiments on OpenLORIS-Scene datasets [36] and our own dataset, evaluating the lifelong capabilities and localization performance under low-textured environments.

\begin{figure}[tpb]	
\centering	
	\includegraphics[width=1.0\linewidth,height=0.8\linewidth]{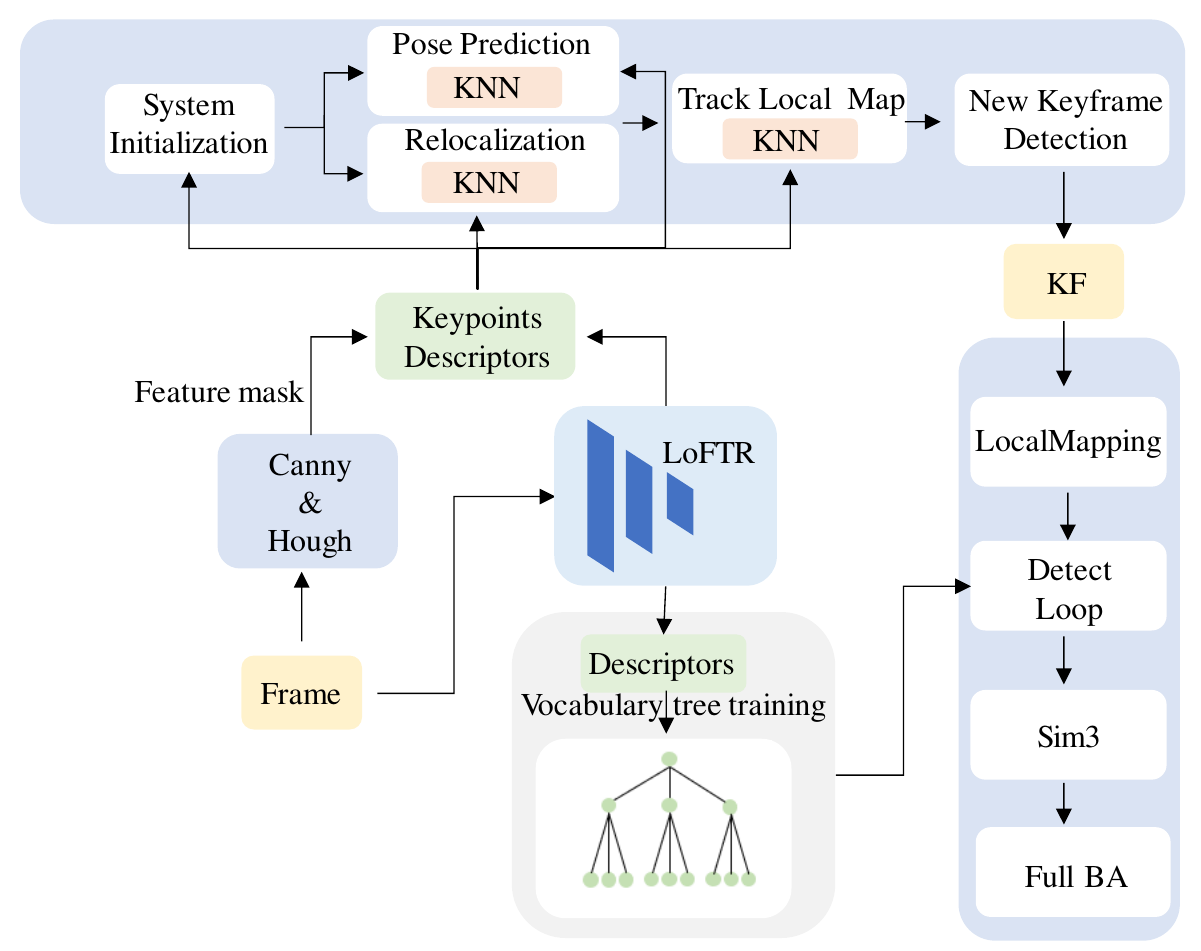}
		\label{framework}
\caption{The system overview of our RWT-SLAM.}	
\end{figure}
We use the pretrained LoFTR model provided by the author to generate the keypoints and descriptors. All experiments are performed on a computer with an intel i7-10700k CPU and NVIDIA 1650 GPU. We sample the images at an interval of 5 frames for the RWT-SLAM system and interpolate the trajectory for the other frames in real time. Our system works at frame rates around 8Hz and can be accelerated with more powerful GPUs.

\begin{figure}[hp]
    \centering
    \subfigure{ 
    \centering
      \includegraphics[width=0.143\textwidth]{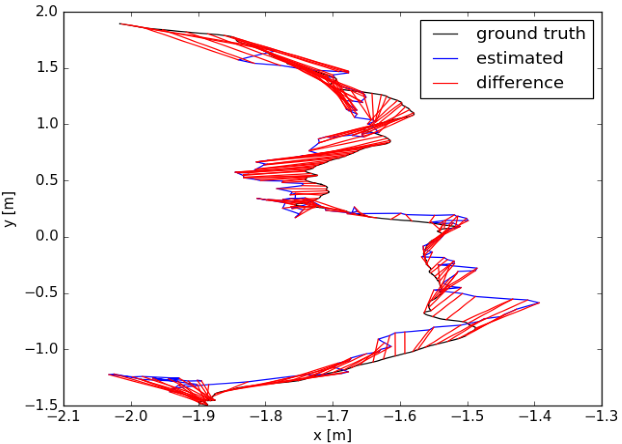}
    }
    \subfigure{   
    \centering
    \includegraphics[width=0.143\textwidth]{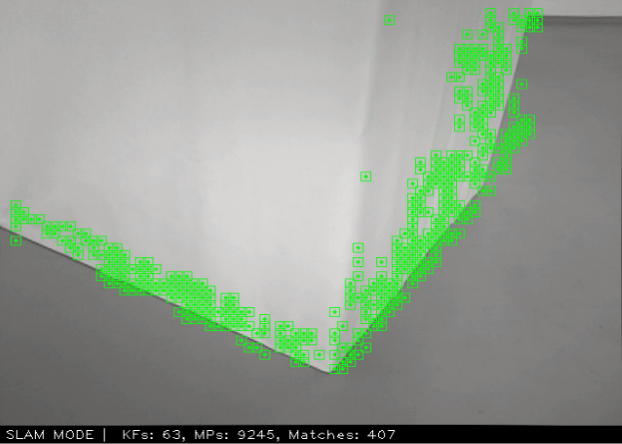}
    }
    \subfigure{   
    \centering
    \includegraphics[width=0.143\textwidth]{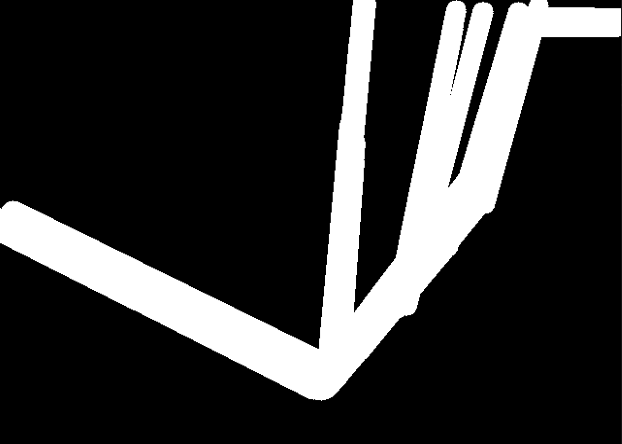}
    }
    \subfigure{
    \centering
          \includegraphics[width=0.143\textwidth]{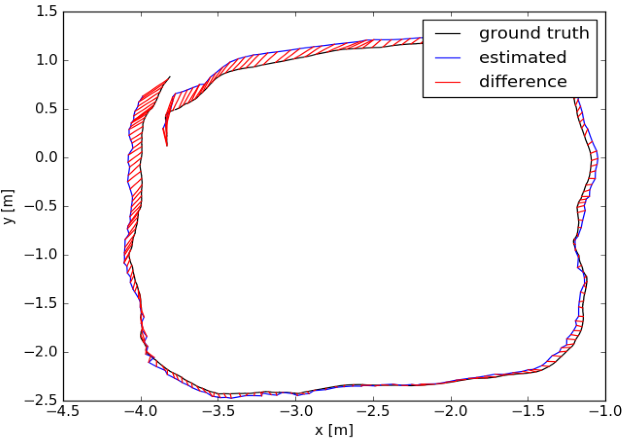}
          }
    \subfigure{   
    \centering
          \includegraphics[width=0.143\textwidth]{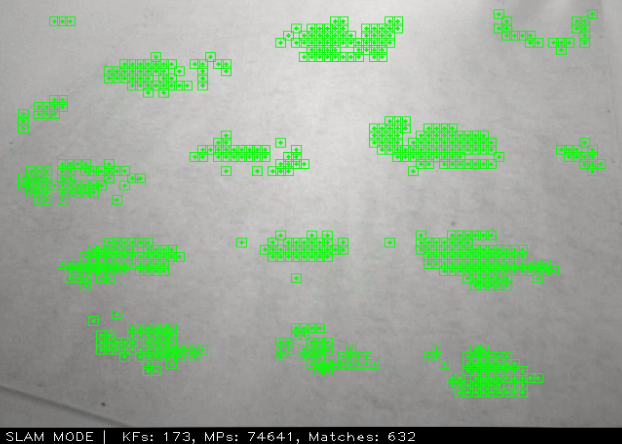}
      }
    \subfigure{   
    \centering
    \includegraphics[width=0.145\textwidth]{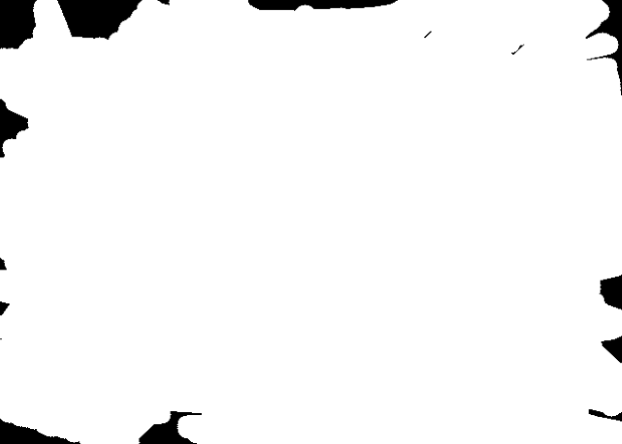}
    }
    \caption{From left to right: The estimated trajectory with error, the tracked keypoints and the feature masks for \textit{fr3_str_notex_near} (top) and \textit{fr3_nostr_notex_near} (below) sequences.}
    % \label{fig:my_label}
\end{figure}

\subsection{Evaluation on TUM RGB-D dataset}
The TUM RGB-D dataset is an excellent dataset with accurate ground truth from motion capture system, which is very suitable for evaluating  the localization accuracy. We select several sequences in \textit{fr3} which have weak or even no-texture for testing.

RWT-SLAM is built upon ORB-SLAM2 [4]. For comparison, we also execute GCN-SLAM [12] and DXSLAM [22], who are both based on ORB-SLAM2 and use learned features at the front-end. Absolute Trajectory Error(ATE) are used for quantitative evaluation. The experiments are conducted with $640\times480$ resolution except for GCN-SLAM, where $320\times160$ resolutions is adopted since this is the best configuration according to their original paper.

\begin{table}[ht]
\caption{RMSE RESULTS ON TUM RGB-D DATASET}
\begin{center}
\begin{tabular}{c|c|c|c||c} 
\hline
sequences               & \begin{tabular}[c]{@{}l@{}}ORB\\SLAM2\end{tabular} & \begin{tabular}[c]{@{}l@{}}GCN\\SLAM\end{tabular} & DXSLAM & \begin{tabular}[c]{@{}l@{}}RWT\\SLAM\end{tabular}  \\ 
\hhline{====:=}
fr3\_cabinet            & -                                                  & \textbf{0.070m}                                    & -      & 0.142m                                              \\ 
\hline
fr3\_str\_notex\_far    & \textbf{0.037m}                                     & 0.072m                                             & -      & 0.044m                                              \\ 
\hline
fr3\_str\_notex\_near   & -                                                  & -                                                 & -      & \textbf{0.212m}                                     \\ 
\hline
fr3\_nostr\_notex\_far  & -                                                  & \textbf{0.086m}                                             & -      & 0.100m                                              \\ 
\hline
fr3\_nostr\_notex\_near & -                                                  & -                                                 & -      & \textbf{0.134m}                                     \\
\hline
\end{tabular}
\end{center}
\end{table}

\begin{figure*}[bp]	
\centering	
	\includegraphics[width=1.0\linewidth,height=0.34\linewidth]{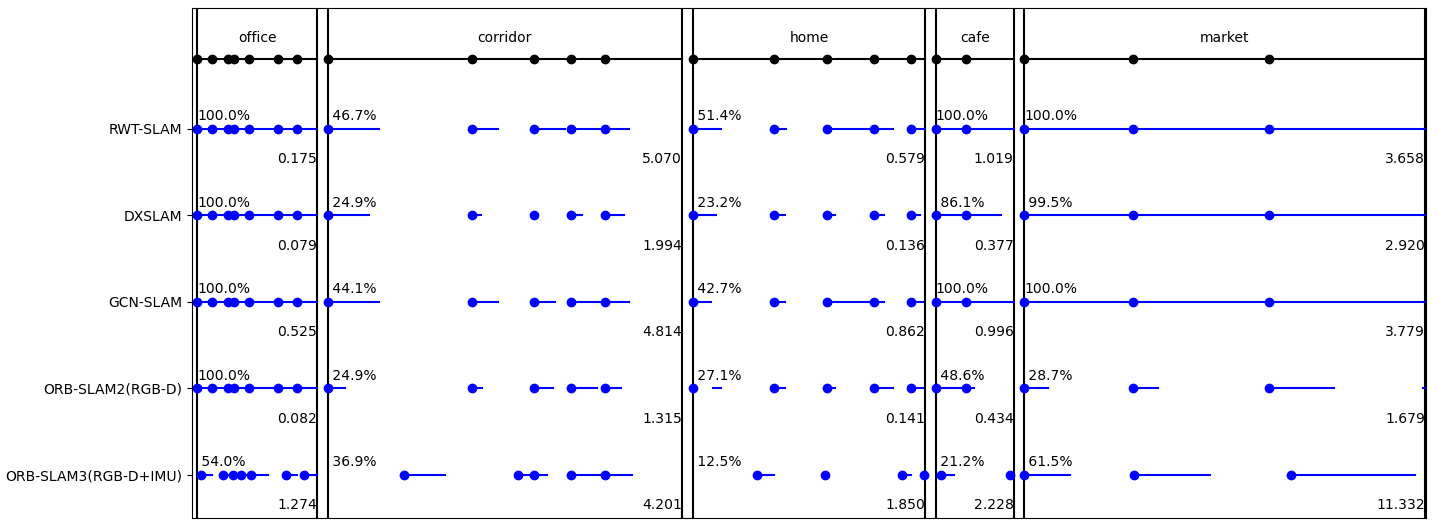}
		\label{OpenLORIS}
\caption{Per-sequence results on OpenLORIS-Scence dataset. Each black dot on the top line indicates the start point of  one sequence in a scene. For each algorithm, blue dots represent successful initialization and blue lines represent the duration of successful tracking. The percentage value on the top left of each scene is the average correct rate $\mathrm{CR}^{\infty}$, with larger meaning more robust. The average ATE RMSE is on the bottom right, with smaller meaning more accurate.}	
\end{figure*}

The comparison results are listed in Table I, where we can see our method succeeds tracking in all sequences. To the best of our knowledge, this is the first visual SLAM algorithm who can survive all of these harsh sequences. For \textit{fr3_cabinet}, there is only a large cabinet in an empty room. Hand crafted keypoints like ORB can be only extracted around the corners or the edges of the cabinet so it is easy to lose tracking. \textit{fr3_str_notex_far} and \textit{fr3_str_notext_near} have some structures but with little texture. The last two sequences are of both little structure and little texture, causing failure for most of the algorithms. Comparing with other learning-based or hand-crafted keypoints, our system are highly reliable in these extreme environments. The trajectory errors, the tracked keypoints as well as the feature masks for two of these sequences are also illustrated in Fig. 4. Meanwhile, we notice that for the sequences that ORB-SLAM2 or GCN-SLAM can survive, RWT-SLAM does not achieve the best accuracy. This may due to the more accurate feature position the ORB or GCN produces once they can survive.

\subsection{Experiment on OpenLORIS Dataset}
There are five scenes in OpenLORIS-Scene dataset, including office, corridor, home, café and market. Some scenes are very challenging because of featureless walls and low illumination, making it impossible to finish the entire tracking for most visual SLAM algorithms. We compare our algorithm with ORB-SLAM and other deep feature based SLAM methods. GCN-SLAM dose not conduct experiment on this dataset and we run the source code of it with the default configurations provided by their paper. Since OpenLORIS provides IMU data, we also test ORB-SLAM3 [37] which is a state-of-the-art visual-inertial SLAM system for comparison. The success rate $\mathrm{CR}^{\infty}$ and the RMSE error of the successful part of the trajectory are used as evaluating metric. The results are shown in Fig. 5, where we can see that RWT-SLAM achieves the longest tracking life on all of these scenes. When the most challenging corridor and home scenes which are highly weak-textured are concerned, our RWT-SLAM outperforms the GCN-SLAM who ranks the second by $5.90\%$ and $20.37\%$  in $\mathrm{CR}^{\infty}$  respectively. It is interesting to find that with the help of IMU, ORB-SLAM3 does not performs better than ORB-SLAM2 in most of the sequences. This may due to the difficulty of the initialization of the IMU in these environments. As to the localization accuracy, our system is comparable to the GCN-SLAM, while inferior than the feature tailored system like ORB-SLAM and DXSLAM. The phenomenon is similar to section IV-A. Nevertheless, our goal is to survive longer and achieve higher robustness under challenging weak-textured scenes. 

\begin{figure*}[ht]
    \centering
    \subfigure[Corridor]{
    \centering
          \includegraphics[width=0.23\textwidth]{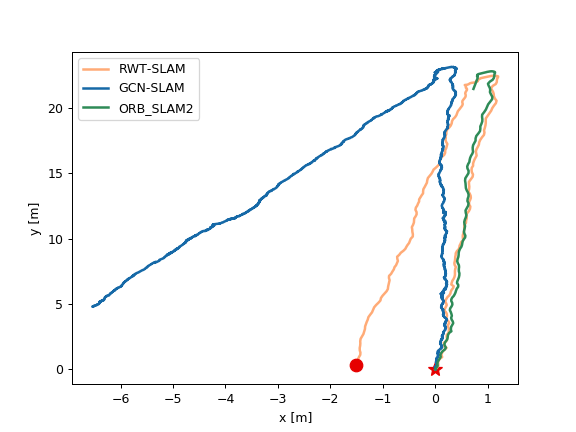}
          }
    \subfigure[Room with wall and floor]{   
    \centering
          \includegraphics[width=0.23\textwidth]{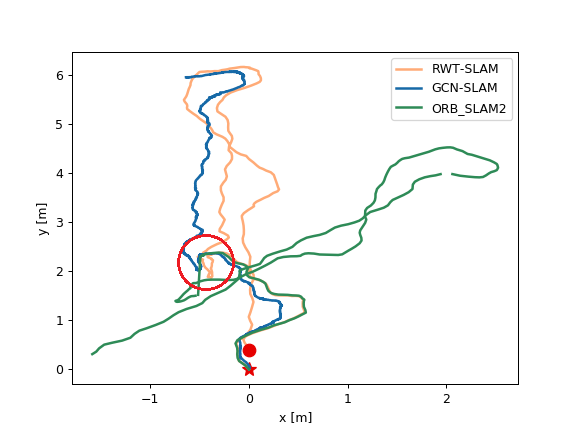}
      }
    \subfigure[Outdoor road]{ 
    \centering
      \includegraphics[width=0.23\textwidth]{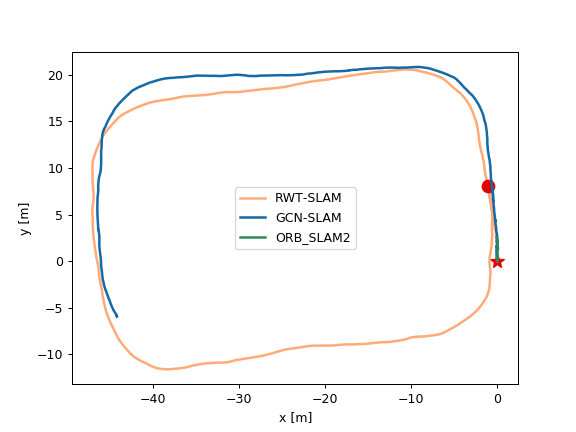}
    }
    \subfigure[Stairs: going up and down]{   
    \centering
    \includegraphics[width=0.23\textwidth]{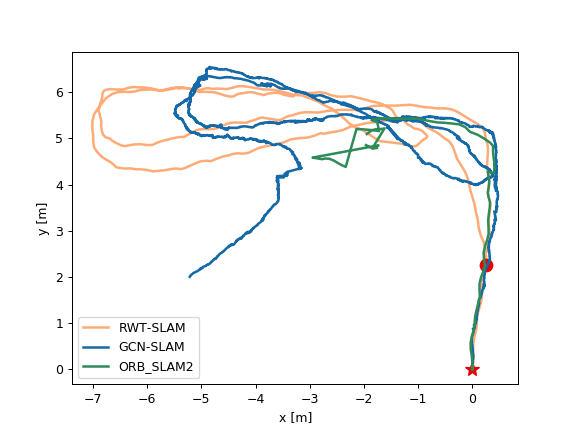}
    }
    \caption{The estimated trajectory for RWT-SLAM, GCN-SLAM and ORB-SLAM2 in our own dataset. We mark the starting and ending points of the trajectory with the star and thick dot respectively. Our method maintains tracking and gives reasonable localization trajectory for all of the sequences while the other two either fail tracking or output 	wrong trajectory.}
\end{figure*}

\begin{figure*}[ht]
    \centering
        \subfigure[Outdoor Road]{
    \centering
          \includegraphics[width=0.46\textwidth]{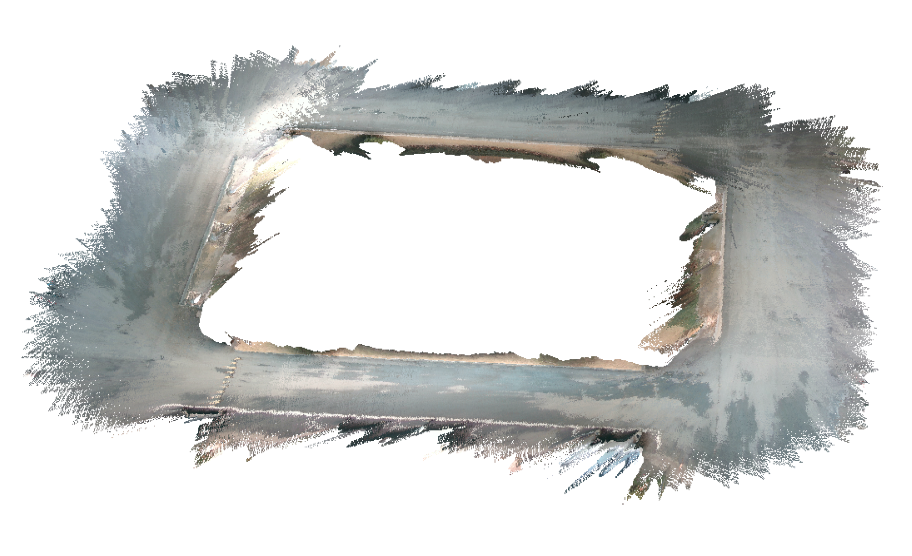}
          }
    \subfigure[Stairs]{   
    \centering
          \includegraphics[width=0.46\textwidth]{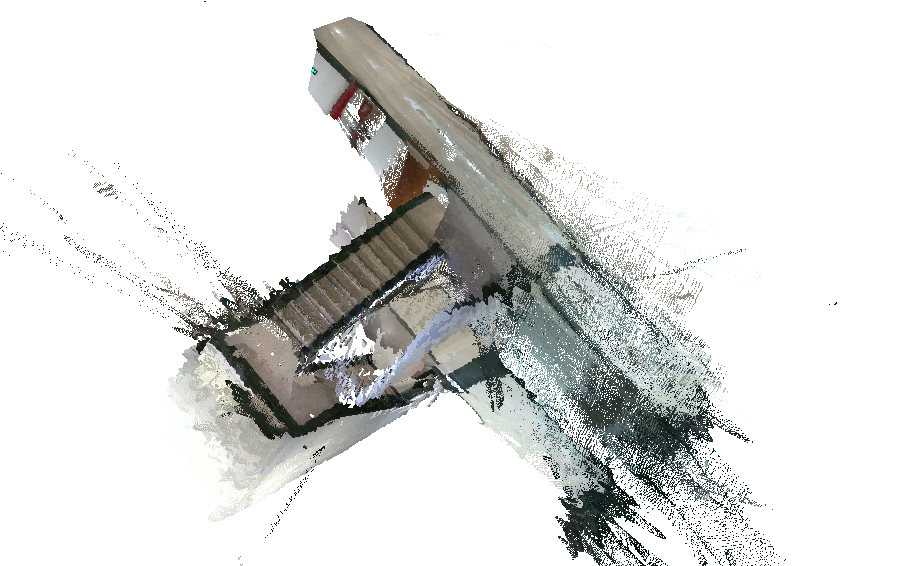}
      }
    \caption{Dense point cloud reconstructions for the Outdoor road and Stairs sequences. The point cloud is obtained by projecting the keyframe’s depth maps with the estimated poses from RWT-SLAM.}
\end{figure*}

\begin{table*}[ht]
\caption{ABLATION STUDIES OF RWT-SLAM.}
\centering
\begin{tabular}{c|c|c|c|c|c|c|c|c|c} 
\hline
         & \begin{tabular}[c]{@{}l@{}}coarse-level\\descriptors\end{tabular} & \begin{tabular}[c]{@{}l@{}}fine-level\\descriptors\end{tabular} & \begin{tabular}[c]{@{}l@{}}line \\mask\end{tabular} & \begin{tabular}[c]{@{}l@{}}KNN \\matcher\end{tabular} & \begin{tabular}[c]{@{}l@{}}fr3\_\\cabinet\end{tabular} & \begin{tabular}[c]{@{}l@{}}fr3\_str\\\_notex\_far\end{tabular} & \begin{tabular}[c]{@{}l@{}}fr3\_str\\\_notex\_near\end{tabular} & \begin{tabular}[c]{@{}l@{}}fr3\_nostr\_\\notex\_far\end{tabular} & \begin{tabular}[c]{@{}l@{}}fr3\_nostr\_\\notex\_near\end{tabular}  \\ 
\hline
(a) & \checkmark & & \checkmark & \checkmark & 0.385m & 0.046m & 0.221m & 0.164m & 0.162m \\
(b) &  & \checkmark & \checkmark & \checkmark & - & 0.450m & - & - & - \\
(c) & \checkmark & \checkmark & & \checkmark & 0.263m & 0.087m & 0.380m & 0.285m & \textbf{0.123m} \\
(d) & \checkmark & \checkmark & \checkmark & & - & 0.074m & 0.216m & - & - \\
RWT-SLAM & \checkmark & \checkmark & \checkmark & \checkmark & \textbf{0.142m} & \textbf{0.044m} & \textbf{0.212m} & \textbf{0.100m} & 0.134m\\
\hline
\end{tabular}
\end{table*}

\subsection{Qualitative Results on our own dataset}
To further demonstrate the performance of RWT-SLAM in real weak-texture areas, we use an Intel RealSense D455 RGB-D camera and collect four sequences under different conditions: a) walking through a corridor, turning 180 degrees at the end of the corridor and going back; b) going into a room looking at the floor and white walls; c) walking on a road outdoor forming a closed loop in the end; d) going upstairs and down back to the start point. Since there is no metric ground truth available, the experiments in this section serves as a complementary testing under real challenging scenes. We compare our system with GCN-SLAM and ORB-SLAM2 and the estimated trajectory results are shown in Fig. 6. As shown in Fig. 6a, ORB-SLAM2 cannot cope with 180 degrees turn and lost tracking at the top right of the trajectory. GCN-SLAM gives wrong rotation prediction and trajectory deviates from the actual waking route while our system keeps reasonable tracking through the sequence. In room sequence, GCN-v2 and ORB features decrease dramatically when encountering walls and floors with weak-texture, leading to incorrect localization stating from the marked red circle shown in Fig. 6b. As shown in Fig. 6c, in outdoor sequence tracking fails immediately for ORB-SLAM2, and GCN-SLAM loses tracking at the third turn. Fig. 6d shows great robustness of RWT-SLAM when dealing with the images containing large featureless areas and motion blurs. Our RWT-SLAM is the only one surviving all of the four challenging sequences with reasonable localization accuracy. The reconstructed map for the outdoor and stair sequences are shown in Fig. 7.

\subsection{Ablation Studies}
In order to demonstrate the effectiveness of each component proposed, we conduct ablation studies on TUM RGB-D dataset. The results of different configurations of RWT-SLAM are shown in Table II. In term of the feature descriptor, RWT-SLAM with coarse-level descriptors only can survive all of the five sequences but with lower localization accuracy, while with fine-level descriptors alone fails in most of the sequences. This is because coarse-level descriptors have large receptive field but with rough resolution, which leads to worse tracking. Fine-level descriptors only captures small receptive filed which is not distinctive enough for matching. Comparing to the full configuration, removing the feature mask module will decrease the localization accuracy. KNN matcher is also essential for the method in the sense that without it only two sequences can be successful treated.

\section{CONCLUSIONS}
In this paper, a robust visual SLAM system based on deep feature is proposed to deal with highly weak-textured environments. Different from the existing detector-based deep features, the state-of-art detector-free network LoFTR is modified to generate dense interest points and the corresponding descriptors. Thanks to the large receptive field provided by the attentional aggregation of the network, we are able to obtain high quality descriptors for scenes with little texture and structure. To make the feature more adaptive to challenging scenes, we present feature masks and retrain the vocabulary for the classical ORB-SLAM framework. KNN matching is also employed during tracking to strengthen the feature matching under extreme environments. Experimental results on various public datasets and our own data prove the success of our system.

\addtolength{\textheight}{-12cm}   % This command serves to balance the column lengths
                                  % sure that you do not shorten the textheight too

\end{document}